# Three Generations of Healthcare IT: From the Digital Record to the Computable Care Process

Alexander Apartsin, PhD[1], Yehudit Aperstein, PhD[2]

[1]School of Computer Science, Faculty of Sciences, Holon Institute of Technology (HIT), 52 Golomb St., Holon 5810201, Israel

[2]Intelligent Systems, Afeka Academic College of Engineering, 218 Bnei Efraim St., Tel-Aviv 6910717, Israel

## Abstract

**Objective.** Healthcare IT is usually organized by the technologies it adopts. We instead organize it by the unit of information a system makes computable, and describe a computational layer whose object is patient-specific clinical intent.

**Approach.** We give criteria for a computational layer, derive three (record, clinical state, and a proposed layer of intent), and formalize the Actionable Clinical Record (ACR) as the third layer's atomic object.

**Discussion.** The framework distinguishes prescribed, observed, and intended process; existing standards represent intent once it is structured but do not recover it from natural communication, the capability we localize. The ACR is complementary to FHIR workflow resources, guidelines, and process mining; a companion feasibility study illustrates tractability for one narrow subproblem.

**Conclusion.** Computable clinical intent is a coherent research direction; the ACR, its readiness ladder, and an executable-correctness evaluation framework are reusable constructs for subsequent work to extend, evaluate, or falsify.

**Keywords:** electronic health records; clinical workflow; actionable clinical record; clinical intent; FHIR

## 1. Introduction

Healthcare information technology is best analyzed through the computational objects it supports rather than the technologies it adopts.[1,2] Organized this way, the field's history is compact: the first generation made the clinical *record* computable, the second the clinical *fact*, and a third, which we propose, makes patient-specific clinical *intent* computable from communication. We use *generation* for the emergence of a new computational layer; the layers accumulate rather than replace one another.

A limitation persists after the first two layers are in place. When a clinician writes "repeat the complete blood count in two weeks" or "refer to cardiology if symptoms persist," the instruction is decisive yet captured only as narrative text, its action, timing, condition, and owner outside the computable substrate. The cost is measurable: test-result follow-up often fails,[3] in one large health system only about one-third of specialist referrals reached a completed visit,[4] and such loop-closure failures, rooted in ambiguous responsibility and care-coordination breakdowns,[5] contribute to diagnostic error and avoidable harm.[6-9]

This Perspective offers a framework and vocabulary rather than a new empirical study: layer criteria and a three-layer organization (Section 2), the Actionable Clinical Record (ACR) as the third layer's atomic object (Section 5), an executable-correctness evaluation framework and companion feasibility study (Section 7), and a research agenda (Section 8).

## 2. A Framework of Computational Layers

We distinguish a computational layer from an incremental trend by five markers: a new atomic *unit of computability*; an *external driver*; an *enabling technology*; a new *class of computation* the unit permits; and a *residual limitation* that motivates the next layer's object. The layers are *cumulative*, each building on those beneath it (Figure 1), and differ in epistemic status: the first two are *retrospective abstractions* over a well-documented history,[1,2] whereas the third is a *prospective hypothesis*, offered as an analytic lens rather than an established periodization. It is complementary to the data–information–knowledge hierarchy[10] and digital-maturity models such as EMRAM,[11,12] which measure adopted capability; our concern is *which* object becomes computable (Table 1).

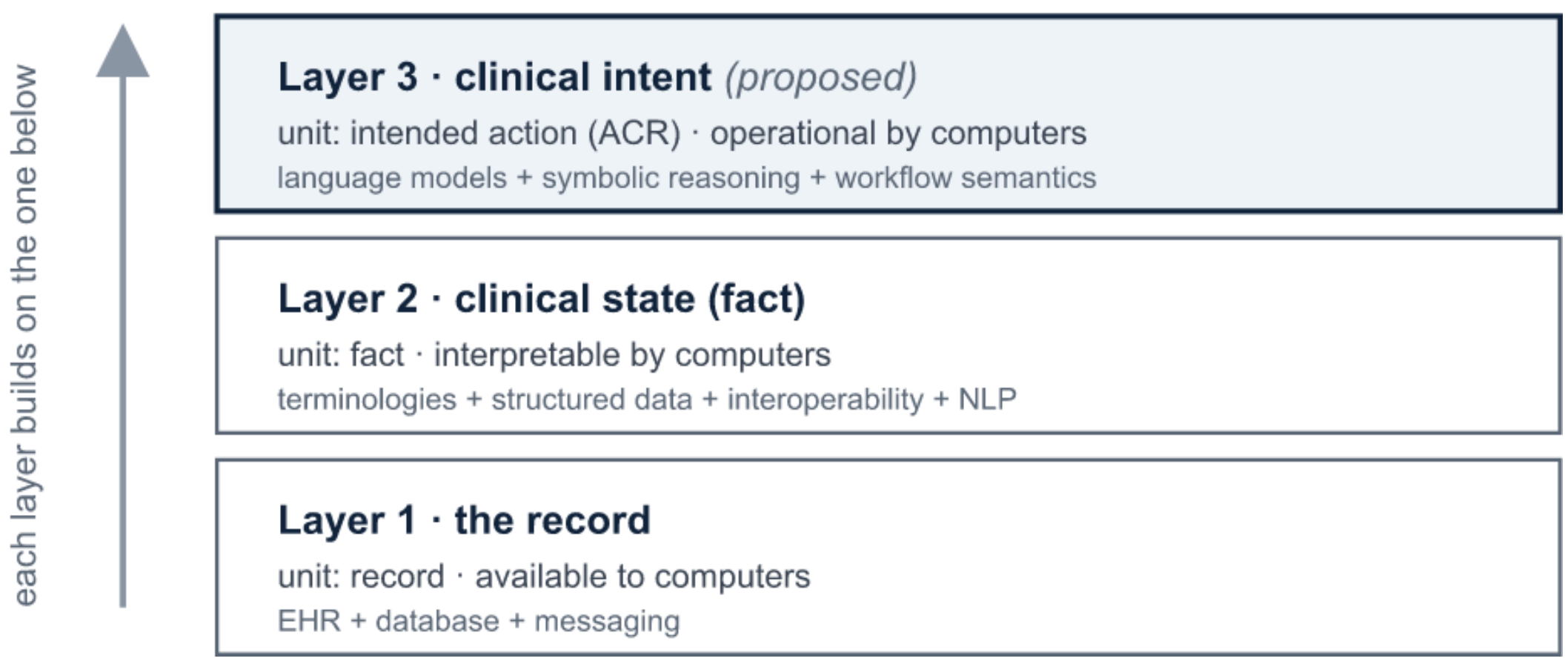

***Figure 1.*** *The three computational layers of healthcare IT, organized by the unit of information each makes computable. Layers accumulate rather than replace; layer 3 is proposed. Alt text: three stacked boxes, record at the base, clinical state above it, and clinical intent at the top, with an upward arrow indicating that each layer builds on the one below.*

| Axis | Layer 1: Record | Layer 2: Clinical state | Layer 3: Clinical intent (proposed) |
|---|---|---|---|
| Unit of computability | Clinical record | Discrete clinical fact | Intended clinical action (ACR) |
| Question answered | What was documented? | What is true about the patient? | What is supposed to happen? |

| External driver | Adoption policy and reimbursement | Interoperability and quality measurement | Care continuity and patient safety |
|---|---|---|---|
| Enabling technology | Transactional databases; messaging | Terminologies; structured data; interoperability; clinical NLP | Language models; symbolic reasoning; workflow semantics |
| Class of computation | Store, retrieve, exchange | Query, reason, predict | Schedule, coordinate, monitor, close |
| Canonical output | Document | Coded fact | Actionable Clinical Record |
| Residual limitation | Meaning remains in narrative | Intent remains in communication | Safe, reliable execution |

***Table 1.*** *The three computational layers against the framework markers. Layers 1 and 2 are retrospective; layer 3 is proposed. The residual limitation of each layer motivates the unit of the next.*

## 3. The First Two Layers: Record and Clinical State

The first layer digitized the clinical record and its transactions, replacing the paper chart with electronic capture, storage, retrieval, transmission, and operations such as order entry and results routing. In the United States, HITECH and Meaningful Use drove rapid nationwide EHR adoption,[13,14] extending the older problem-oriented-record ideal.[15] Its enabling technology, the transactional database, stored and moved information without making its clinical meaning computable across systems.

The second layer turned the record into computable clinical *state*. Its enabling stack was broad, not a single technique: structured data entry, controlled terminologies,[16] computerized order entry, data warehouses, and interoperability standards[17,18] mattered as much as language processing (MedLEE, cTAKES).[19,20] These enabled the first broad wave of computation *on* clinical data: decision support, quality measurement, registries, and research.[21] A tension remained: encoding meaning into structured fields captures facts, while much clinical reasoning and many intended actions remain expressed only in narrative.[22]

## 4. The Remaining Computational Gap

The first two layers made clinical *state* computable, but many patient-specific *intended actions* remain non-computable while they are expressed only in unstructured communication. Healthcare IT can represent process, but only once a person or system has structured it: FHIR can represent patient-specific intent through *CarePlan*, *ServiceRequest*, and *Task*, while *PlanDefinition* defines reusable protocols that are instantiated per patient;[23] computer-interpretable guidelines and care-pathway models encode executable protocols;[24-26] and process mining reconstructs pathways from event logs.[27,28] These paradigms are complementary and often provide infrastructure to operationalize the proposed layer's outputs. The gap is isolated by distinguishing three kinds of process information (Table 2): *prescribed* (what should generally happen), the domain of guidelines and decision support; *observed* (what actually happened), the domain of records and process mining; and *intended* (what is supposed to happen for a specific patient), often expressed in communication rather than structured directly. Existing workflow paradigms represent and operationalize intended process once it has been structured, but they do not themselves specify recovery from unstructured communication; the third layer concerns that recovery and its conversion into executable form.

| Process type | Meaning | Typical source | Existing paradigm |
|---|---|---|---|
| Prescribed | What should generally happen | guideline / protocol | computer-interpretable guidelines; decision support |
| Observed | What actually happened | event log / EHR | process mining |
| Intended | What is supposed to happen for a specific patient | clinical communication | represented once structured (FHIR); recovery from communication proposed here |

***Table 2.*** *Three kinds of process information. The categories describe the semantic status of a process statement, not exclusive source types. The proposed third layer targets recovery of patient-specific intended process that remains expressed in natural communication.*

## 5. Computable Clinical Intent

We propose that the third layer makes patient-specific clinical intent computable by recovering it from communication as a structured, executable representation, using one terminology throughout: clinical *communication* may express *intent*; an intent decomposes into intended *actions*; each intended action is represented by an *Actionable Clinical Record*; a set of ACRs and related events is a clinical *workflow*; and the enacted sequence over time is the *care process*.

### 5.1 The Actionable Clinical Record

Definition 1 (Actionable Clinical Record). An Actionable Clinical Record is a source-grounded representation of a patient-specific intended clinical action, with the semantic attributes required to interpret, execute, monitor, or audit it: the tuple

ACR = ⟨ action, target, actor, temporal constraint, condition, dependency, status, provenance, confidence ⟩

in which *action*, *status*, and *provenance* are required; *target*, *actor*, *temporal constraint*, *condition*, and *dependency* are populated when communicated or explicitly inferred; and *confidence* is a per-attribute map recording calibrated recovery-system uncertainty for inferred attributes.

Each attribute has a fixed semantics (Figure 2): *action*, the requested operation; *target*, the object acted upon, linked to a layer-2 coded concept where possible; *actor*, the responsible performer rather than the speaker; *temporal constraint*, a normalized time (absolute, relative, event-anchored, or recurring) rather than a date alone; *condition*, a proposition governing applicability; *dependency*, ordering and prerequisites; *status*, the operational lifecycle state; and *provenance*, the source span, speaker, and encounter.

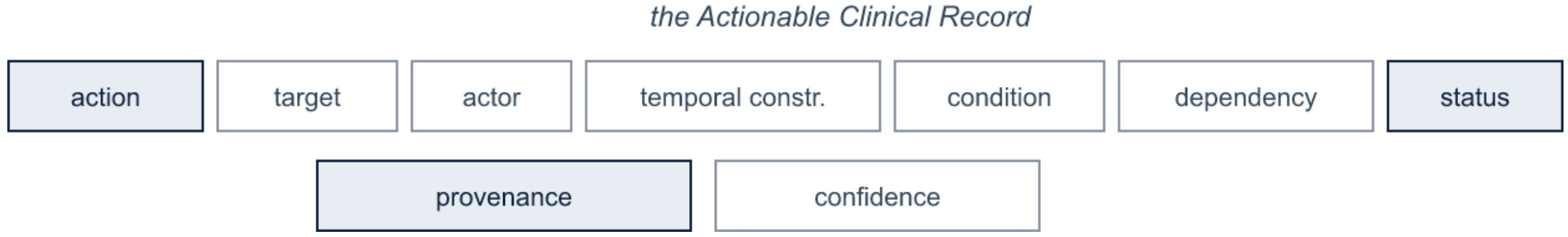

***Figure 2.*** *The ACR schema. Alt text: nine labeled boxes forming the ACR tuple; action, status, and provenance are shaded to mark them as required; target, actor, temporal constraint, condition, dependency, and confidence are outlined as optional.*

The third layer spans two distinct axes that together prevent over- and underclaiming. A representational *readiness* ladder runs *mentioned* → *interpreted* → *actionable* → *executable*: recognizing "complete blood count" in a note is layer 2; extracting "repeat complete blood count" is partial; adding "two weeks" and a responsible actor makes it actionable; resolving a due date and workflow object makes it executable. Separately, the ACR's *status* attribute records the operational lifecycle of the recovered action, from *recovered* and *confirmed* to *active* and then *completed* or *cancelled*, without changing its representational readiness; detecting completion or escalation closes the loop.

### 5.2 From communication to workflow

The third layer is a bridge rather than a replacement electronic health record (Figure 3). Its inputs are the range of clinical communication; its output is a set of ACRs that maps onto infrastructure the field already has.[17,23] The mapping situates the ACR *upstream* of these resources rather than as a competitor, translating attributes into workflow elements according to the target resource and profile: for example, *action* and *target* into a *Task* or *ServiceRequest*, *temporal constraint* into the resource-appropriate timing element, *actor* into performer or owner semantics, and *dependency* and *status* into resource-appropriate relationship and lifecycle fields (*basedOn* or *partOf* where they apply, orchestration constructs for prerequisites) whose FHIR semantics differ from the ACR's and need profile-specific mapping. Recovery *provenance* can itself be represented with a FHIR *Provenance* resource or profile, while *confidence* generally remains metadata of the recovery system.

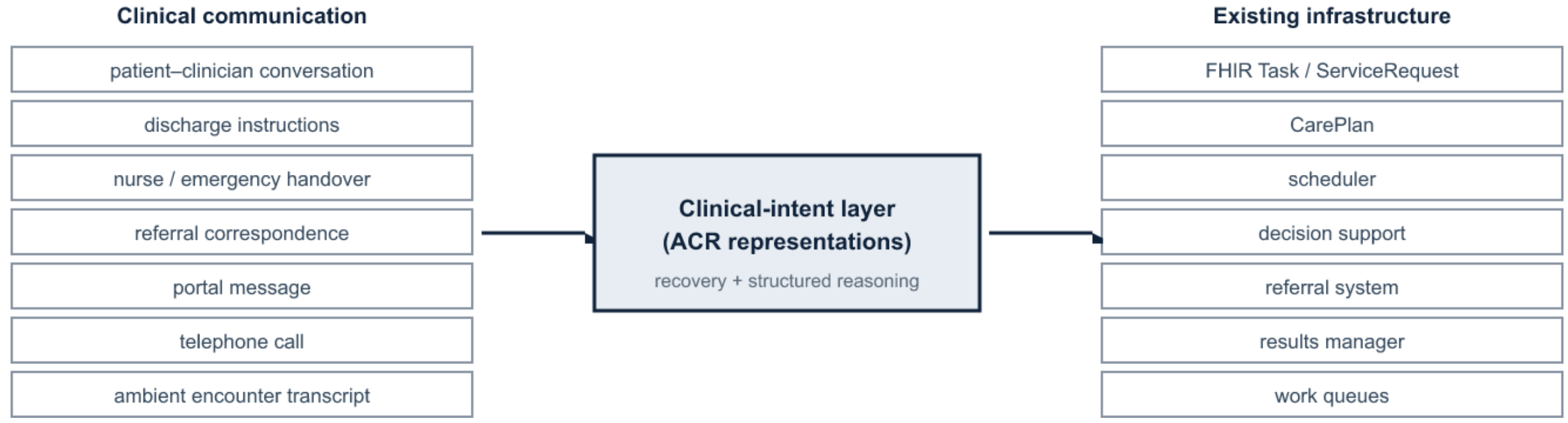

***Figure 3.*** *ACRs as the bridge between clinical communication and existing workflow infrastructure. Alt text: communication sources on the left feed a central clinical-intent layer containing ACR representations, which maps to existing infrastructure such as FHIR resources, schedulers, and referral and results systems on the right.*

Follow-up is a well-documented case, one of a broader class of communication-derived process objects sharing the ACR representation: conditional escalation, medication transitions, handoffs, referrals, and pending-result responsibility. Recent shared tasks extract structured orders directly from clinical conversations,[29] while ambient documentation shows these conversations are already captured.[30]

### 5.3 Relation to existing paradigms

Two mature lines can be mistaken for the third layer. Clinical NLP already extracts concepts, relations, events, temporality, and follow-up recommendations, increasingly with language models,[19,20,31-36] and directly extracts physician action items and medical decisions from clinical notes.[37,38] Even order extraction from conversation targets a flat schema scored by text overlap;[29] the distinction the ACR draws is operational, requiring that a recovered action be source-grounded, temporally normalized, and carry the actor, condition, and dependency structure needed to execute rather than only index it. Computer-interpretable guidelines encode generic, population-level protocols;[24,25] the third layer recovers patient-specific records from a particular encounter, capturing what a clinician instructed for this patient.

## 6. Reliability and Architecture

A more capable generator does not remove the third layer's requirements: auditability, verifiable temporal correctness, and calibrated deferral are properties of the output contract, not model capability. Large language models show clinical information-extraction and medical-reasoning capability[35,39] but can produce unreliable or unsupported outputs;[40] this matters most in binding an action to its time, a setting where general-purpose temporal reasoning is known to be weak.[41]

These requirements motivate a hybrid decomposition in which learned components interpret and explicit components enforce verifiable structure.[42,43] Because some workflow semantics (calendar arithmetic, dependency ordering, status transitions) are formally specified, learned extraction supplies candidate actions and arguments[44] while deterministic reasoning resolves times and due dates[45-48] and selective prediction routes uncertain cases to a clinician. The requirements are invariants: every acted-upon record traces to a source span, speaker, and encounter;[49] deterministic semantics are computed, not generated; the system may abstain; higher-consequence actions receive tighter human review;[50] an immutable trace links communication to action for audit; and extracted content stays distinct from inferred. Human confirmation is the default until task-specific calibration and prospective safety evidence justify higher automation.

## 7. Evaluation Framework and Feasibility

Because third-layer outputs can drive clinical action, they should be evaluated on *executable correctness* rather than text overlap. We propose an evaluation framework for ACR recovery with measures reported separately: action detection; argument extraction; action–time, condition, and actor linking; temporal-normalization error; whole-record exact match; unsupported- and omitted-action rates; calibration and selective-risk behavior; and source-provenance accuracy. These assess ACR recovery; loop-closure outcome, a downstream endpoint shaped by workflow rather than recovery correctness, is reported separately. This framework, not any single score, is the intended contribution.

A companion study offers a formative feasibility test of one minimal task, follow-up-instruction extraction, evaluating a hybrid neural-symbolic pipeline on a controlled benchmark against generative baselines.[51] On the action–time pairing the executable representation requires, the hybrid pipeline substantially outperforms general-purpose language models, which recognize actions but bind them to times unreliably. This is a controlled feasibility demonstration for one narrow subproblem, not evidence for the framework, whose value rests on its constructs.

## 8. Research Agenda

The framework defines a program in four domains. **Representation:** a consensus ACR ontology, with attribute semantics, constraint types, provenance, and reconciliation when communications revise or revoke a plan. **Inference and reliability:** ACR recovery from multilingual and multimodal communication, temporal reasoning, cross-message reconciliation, calibrated abstention, distinguishing patient from clinician intent, and separating who spoke from who is responsible for acting. **Evaluation:** authentic-data benchmarks, the framework of Section 7 across tasks, and cross-institution and cross-language generalization. **Integration and impact:** ACR-to-FHIR write-back into scheduling, results-management, and referral systems,[23] review-queue design, regulatory classification, and whether closed-loop operation improves outcomes.

Moving from synthetic corpora to de-identified real-world notes is the central empirical risk; safe deployment presupposes the invariants of Section 6. The proposal makes testable predictions: machine-actionable use of FHIR workflow resources should grow relative to free-text follow-up; evaluation should shift toward executable-correctness measures; reliable systems should expose source-linked, auditable records with selective review; and ambient documentation should extend from notes to tasks and orders.[30,52,53] More decisively, two independent claims are falsifiable. The ACR representation is redundant if direct mapping into existing workflow standards, with generic recovery metadata, captures the required action, actor, temporal, conditional, dependency, provenance, and uncertainty semantics without ACR-specific distinctions. The evaluation framework is redundant if standard extraction metrics discriminate operationally unsafe outputs as well as the executable-correctness measures across task classes.

### 9. Conclusion

Healthcare IT has made the record and then the clinical fact computable; we propose patient-specific clinical intent as the next layer. What characterizes it is not generating clinical text but recovering, from ordinary communication, source-grounded verifiable representations of intended action, formalized as the Actionable Clinical Record upstream of existing workflow infrastructure. We offer the framework, the ACR, and its evaluation framework as reusable constructs to operationalize, evaluate, extend, or falsify.

### Data Availability

This article is a conceptual Perspective; no new data were generated or analyzed, and it introduces no new datasets. The illustrative feasibility results summarized in Section 7 are drawn from a companion study;[51] the controlled benchmark and evaluation described there, including its own data-availability terms, are reported in that study, and no data were newly collected for the present article. The constructs proposed here, the Actionable Clinical Record schema (Section 5) and the executable-correctness evaluation framework (Section 7), are fully specified in the text and require no separate data deposit. No custom code was produced for this article.

### Author Contributions (CRediT)

A. Apartsin: Conceptualization, Methodology, Writing – original draft. Y. Aperstein: Conceptualization, Methodology, Writing – review and editing.

### Conflicts of Interest

None declared.